\documentclass{article} 
\usepackage{iclr2026_conference,times}

\usepackage{amsmath,amsfonts,bm}

\def\eqref#1{equation~\ref{#1}}

\def\1{\bm{1}}

\def\vb{{\bm{b}}}

\def\ve{{\bm{e}}}

\def\vh{{\bm{h}}}

\def\vk{{\bm{k}}}

\def\vn{{\bm{n}}}
\def\vo{{\bm{o}}}

\def\vq{{\bm{q}}}

\def\vv{{\bm{v}}}
\def\vw{{\bm{w}}}
\def\vx{{\bm{x}}}
\def\vy{{\bm{y}}}

\def\mA{{\bm{A}}}

\def\mC{{\bm{C}}}
\def\mD{{\bm{D}}}

\def\mI{{\bm{I}}}

\def\mO{{\bm{O}}}

\def\mW{{\bm{W}}}

\DeclareMathAlphabet{\mathsfit}{\encodingdefault}{\sfdefault}{m}{sl}
\SetMathAlphabet{\mathsfit}{bold}{\encodingdefault}{\sfdefault}{bx}{n}

\def\sK{{\mathbb{K}}}

\def\sR{{\mathbb{R}}}

\def\sV{{\mathbb{V}}}

\usepackage{hyperref}
\usepackage{url}
\usepackage{booktabs}
\usepackage[textwidth=1.1in]{todonotes}
\usepackage{cleveref}
\AddToHook{cmd/appendix/before}{%
    \crefalias{section}{appendix}%
    \crefalias{subsection}{appendix}
    \crefalias{subsubsection}{appendix}
}
\usepackage{subcaption} 
\usepackage{amsthm}
\usepackage{wrapfig}

\theoremstyle{definition}

\newcommand{\modelname}[1]{gLSTM}
\newcommand{\osq}[1]{over-squashing}
\newcommand{\Osq}[1]{Over-squashing}
\newcommand{\OSq}[1]{Over-Squashing}
\newcommand{\layer}[1]{l}
\newcommand{\Layer}[1]{L}

\definecolor{alvaroblue}{HTML}{0065a9}

\title{\modelname{}: Mitigating \OSq{} by \\ Increasing Storage Capacity}

\author{
Hugh Blayney$^{1}$,
\'Alvaro Arroyo$^{1}$,
Xiaowen Dong$^{1}$,
Michael M. Bronstein$^{1,2}$ \vspace{0.2em} \\
\hspace{0.3em}${}^{1}$University of Oxford \\
\hspace{0.3em}${}^{2}$AITHYRA \vspace{0.2em}\\
\hspace{0.3em}\texttt{hugh@robots.ox.ac.uk}
}

\arxivcopy
\renewcommand{\headrulewidth}{0pt}
\begin{document}

\maketitle

\begin{abstract}
Graph Neural Networks (GNNs) 
leverage the graph structure to transmit information between nodes, typically through the
message-passing mechanism. While these models have found a wide variety of applications, they are known to suffer from {\em \osq{}}, where information from a large receptive field of node representations is collapsed into a single fixed sized vector, resulting in an information bottleneck. In this paper, we re-examine the \osq{} phenomenon through the lens of model \emph{storage and retrieval capacity}, which we define as the amount of information that can be stored in a node's representation for later use. We study some of the limitations of existing tasks used to measure \osq{} and introduce a new synthetic task to demonstrate that an information bottleneck can saturate this capacity. Furthermore, we adapt ideas from the sequence modeling literature on associative memories, fast weight programmers, and the xLSTM model to develop a novel GNN architecture with improved capacity. We demonstrate strong performance of this architecture both on our capacity synthetic task, as well as a range of real-world graph benchmarks.
\end{abstract}

\section{Introduction}
Graph Neural Networks (GNNs) \citep{sperduti1993encoding, gori2005new, scarselli2008graph, NN4G, bruna2014spectral, defferrard2017convolutional} have emerged as a standard framework for learning on graph-structured data. The majority of these models follow a \textit{message passing} paradigm, where nodes iteratively exchange information with neighbors, commonly referred to as Message-Passing Neural Networks (MPNNs). Examples of this family of architectures include GCN \citep{kipf2017semisupervised}, GAT \citep{Velickovic2018GraphAN}, GIN \citep{xu2018powerful}, and GraphSAGE \citep{hamilton2017inductive}.

Since each MPNN layer exchanges information between neighboring nodes to update node representations, the number of layers thus dictates the receptive field: the set of nodes over which information is aggregated. Deep MPNNs are, in theory, desirable as they can model long-range dependencies, but scaling to many layers has historically been difficult due to two pervasive issues that have received significant attention in the literature: \textit{over-smoothing} and \textit{\osq{}}. We focus on the latter in this work.

\Osq{} was initially identified by \citet{Alon2020OnTB} as a problem of compressing information from a node's receptive field into a fixed-size vector. This was linked with depth and long-range dependencies, since receptive fields tend to grow exponentially with depth. Later work \citep{Topping2021UnderstandingOA,Giovanni2023OnOI} identified that this bottleneck could also result in low sensitivity as measured by the node Jacobian, linking graph topology and aspects of model architecture via an upper bound on this Jacobian. This low sensitivity arises due to repeated degree normalization and application of a contractive nonlinearity over many layers. \citet{ArnaizRodrguez2025OversmoothingH} suggest that these two descriptions of \osq{} are not the same, and that \citet{Alon2020OnTB} define it as a problem of computational graph bottlenecks, while later work often defines it as a problem of topological bottlenecks. We discuss this separation and its relation to our work in \Cref{appen:only_computational_tree}.

Instead of separating by issues of computational tree structure and bottlenecks, we suggest an alternative separation by resultant \emph{failure mode}: limited information storage capacity, and low sensitivity. In light of this, we highlight another divergence in the literature: the work of \citet{Alon2020OnTB} implicitly described \osq{} as a capacity problem, and later work re-framed it as a problem of sensitivity. Focusing on the failure modes themselves not only allows us to revisit the issue of capacity originally discussed in \citet{Alon2020OnTB}, but also to more directly motivate benchmarks and remedies, both of which we discuss in this paper. We further contextualize this issue of capacity by studying it in \emph{isolation} from sensitivity issues. We believe this focus not only provides a more complete understanding of \osq{} but also highlights new directions to mitigate it.

To combat \osq{}, existing research has focused on ameliorating topological bottlenecks through \emph{rewiring} \citep{gasteiger2019diffusion,gutteridge2023drew,nguyen2023revisiting} and controlling the flow of information \citep{Bresson2017ResidualGG,finkelshtein2024cooperative,errica2025adaptive} -- targeting topological and general computational bottlenecks respectively. However, these bottlenecks are only an issue if they harm performance in some way: in this work we discuss issues of (1) reduced sensitivity and (2) saturated storage capacity.
{Our proposed architecture in \Cref{sec:architecture} targets the latter failure mode: adapting the MPNN architecture to improve its \textit{ability to store and retrieve information}. Framing \osq{} as a capacity limitation that can be addressed at the architecture level exposes a previously unexplored path, and our results validate this direction.}

To improve MPNN storage capacity we turn to the sequence modeling literature, which has a long history of tackling equivalent problems \citep{Hochreiter1997LongSM, orvieto2023resurrecting, gu2023mamba,Beck2024xLSTMEL,Arora2023ZoologyMA}. Taking inspiration from these works, we introduce an MPNN architecture that utilizes associative memory \citep{Beck2024xLSTMEL,Schlag2021LinearTA,Hopfield1982NeuralNA}, and demonstrate that this exhibits improved storage capacity.

\paragraph{Contributions.}
Our main contributions are as follows. In \Cref{sec:osq_as_capacity}, we \textbf{re-characterize \osq{} into two distinct failure modes}: {\em saturating capacity} and {\em low sensitivity}, which we term capacity \osq{} and sensitivity \osq{} respectively. We discuss in \Cref{sec:existing_osq_limitations} the pitfalls of widely used \osq{} tasks, which either fail to evaluate capacity at all, or evaluate the two issues in tandem and are thus unable to separate their effects. In \Cref{sec:neighbor-recall-explanation}, we introduce a novel synthetic task, which to our knowledge is the first that {\bf measures capacity \osq{} in isolation}. In \Cref{sec:architecture}, we present a new MPNN architecture based on the recent xLSTM architecture \citep{Beck2024xLSTMEL}, which uses \textbf{associative memory to increase capacity}, explicitly targeting this capacity \osq{} viewpoint. \Cref{sec:experiments} demonstrates that this architecture performs well on our synthetic capacity task and a range of real-world benchmarks, and \Cref{sec:nar_sensitivity} demonstrates empirically that \textbf{capacity \osq{} can occur separately from sensitivity \osq{}}.

\section{Background and Related Work}\label{sec:am_background}

\paragraph{Message Passing Neural Networks}

Let a graph $\mathcal{G}$ be a tuple $(\mathcal{V},\mathcal{E})$ where $\mathcal{V}$ is the set of nodes and $\mathcal{E}$ the set of edges.  
An edge from node $u$ to $v$ is denoted $(u,v)\in\mathcal{E}$.  
The connectivity is encoded by the adjacency matrix $\mA\in\mathbb{R}^{|\mathcal{V}|\times |\mathcal{V}|}$, where $\mA_{uv}=1$ if $(u,v)\in\mathcal{E}$ and $0$ otherwise.  
Each node $v$ has a feature vector $\vx_v\in\mathbb{R}^d$.

GNNs are functions  
$f_{\boldsymbol{\theta}}\!:\!(\mathcal{G},\{\vx_v\})\mapsto \vy$ with parameters $\boldsymbol{\theta}$, trained via gradient descent to predict node- or graph-level labels $\vy$.  
These models typically take the form of MPNNs, which compute latent representations by composing $\Layer{}$ layers of the following node-wise operation:
\begin{equation}
\vh_{u}^{(\layer{})}=\phi^{(\layer{})}\!\big(\vh_{u}^{(\layer{}-1)},\;
      \psi^{(\layer{})}(\{\vh_{v}^{(\layer{}-1)}:(u,v)\in\mathcal{E}\})\big),
\end{equation}
where $\psi^{(\layer{})}$ is a permutation-invariant \emph{aggregator}, $\phi^{(\layer{})}$ combines neighbor messages with the previous embedding and $\vh_v^{(0)} = \vx_v$. Throughout, we use “GNN’’ and “MPNN’’ interchangeably. Note we depart from the more usual notation of $k$ for layer index to avoid confusion with \emph{keys}, introduced in \Cref{sec:neighbor-recall-explanation}. The most commonly used aggregation function takes the form
\begin{equation}\label{eq:mpnn_aggr}
\psi^{(\layer{})}(\{\vh_{v}^{(\layer{}-1)}:(u,v)\in\mathcal{E}\})
   =\sum_{v}\mO_{uv}\,\vh_{v}^{(\layer{}-1)},
\end{equation}
where $\mO\in\mathbb{R}^{|\mathcal{V}|\times |\mathcal{V}|}$ is some message-passing matrix. For GCN \citep{kipf2017semisupervised}, $\mO = \Tilde{\mD}^{-1/2}\Tilde{\mA}\Tilde{\mD}^{-1/2}$ with $\Tilde{\mA} = \mA + \mI$ for diagonal $\Tilde{\mD}\in\mathbb{R}^{|\mathcal{V}|\times |\mathcal{V}|}$ with $\Tilde{\mD}_{ii} = \sum_j \Tilde{\mA}_{ij}$. We frequently denote the set of message-passing neighbors of node $u$ as $\mathcal{N}_u = \{ v \in \mathcal{V} \mid \mO_{uv} \neq 0 \}$ -- if the message-passing matrix is layer-dependent, we may superscript this with a layer index.

\vspace{-0.3cm}
\paragraph{Hopfield Networks}

Hopfield networks \citep{Hopfield1982NeuralNA} were introduced as associative memories storing binary patterns via a ``Hebbian learning'' rule in which patterns are directly encoded via an outer product. Later modifications increased storage capacity \citep{krotov2016dense,krotov2018dense} or adapted to continuous states \citep{hopfield1984neurons,koiran1994dynamics}. Retrieval is typically a multi-step iterative process; \citet{Ramsauer2020HopfieldNI} introduced a variant with the ability to retrieve patterns in a single step, demonstrating equivalence with Transformer \citep{vaswani2017attention} key-value recall.

\vspace{-0.3cm}
\paragraph{Fast Weight Programmers}

Fast Weight Programmers (FWPs) are a class of neural network motivated by the idea of allowing variable network weights dependent on the input - termed \emph{fast weights}. One method to ``program'' the fast weights is to take outer products of learned projections of the input \citep{Schmidhuber1992LearningTC}. \citet{Schlag2021LinearTA} observe that -- up to normalization and activation function differences -- linear Transformers \citep{Katharopoulos2020TransformersAR} are equivalent to FWPs.

\vspace{-0.3cm}

\paragraph{xLSTM: Associative Memory for Language Modeling}

Recent work \citet{Beck2024xLSTMEL,Beck2025xLSTM7A} introduced xLSTM, a development of the original LSTM \cite{Hochreiter1997LongSM} architecture that resulted in a performant recurrent neural network capable of language modeling. Of relevance to our work are the following limitations that xLSTM aims to address: the inability to ``revise storage decisions'' and the limited storage capacity of the scalar cell states. The first of these is addressed through modifying the original LSTM gating to use exponential activation functions. The second is addressed by introducing associative memory, updated using an outer product update rule equivalent to that of FWPs to store keys and values (see \Cref{appen:xlstm} for more details).

\section{The Two Failure Modes of \OSq{}}\label{sec:osq_as_capacity}

\Osq{} was initially introduced by \citet{Alon2020OnTB} as an issue of \textbf{storage capacity}. They observed that recurrent sequence models exhibit a bottleneck in representing all the information from their past inputs, and this bottleneck exists in a more harmful form in GNNs, in which the information receptive field grows exponentially. They introduced a synthetic task to measure \osq{} by propagating information through various sizes of binary tree.

Later research identified that this computational graph bottleneck \emph{also} resulted in \textbf{low sensitivity} and issues of signal propagation. \citet{Topping2021UnderstandingOA,Giovanni2023OnOI} quantified this low sensitivity via the Jacobian of node representations, establishing the following sensitivity bound: for an MPNN with ${\layer{}}$ layers, $c_{\sigma}$ Lipschitz constant of the activation, $w$ maximal entry-value over weight matrices, $d$ embedding dimension and $u,v \in \mathcal{V}$, one has 
\vspace{-0.3cm}
\begin{equation}\label{eq:sens_bound}
    \left \lVert \frac{\partial \vh_v^{(\layer{})}}{\partial \vh_u^{(0)}} \right  \rVert_{1} \leq \underbrace{\left ( c_{\sigma} wd \right )^{\layer{}}}_{\text{model}} \overbrace{\left ( \mO^{\layer{}} \right )_{uv}}^{\text{topology}},
\end{equation}
where $\mO$ is the message passing matrix used by the MPNN as in \Cref{eq:mpnn_aggr}. This bound establishes that low sensitivity 
results from both graph topology as well as factors intrinsic to the MPNN model. In particular, sensitivity is lowered by the nature of the message-passing, where the culprit is successive powers of a degree-normalized adjacency matrix. It is also lowered by the contractive nature of the nonlinearity $\sigma$ and the values of the weight matrices, as established in \citep{Arroyo2025OnVG}. Despite this analysis being purely one of sensitivity rather than capacity, it was also termed \osq{}, 
and has been successful in establishing links to other areas, including the expressive power of MPNNs \citep{Giovanni2023HowDO} and graph effective resistance \citep{black2023understanding}.

We argue that there are two \emph{distinct} problems arising from bottlenecks in MPNNs: {\em reduced sensitivity} (sensitivity \osq{}) and {\em saturating storage capacity} (capacity \osq{}). Due to the influential paper of \citet{Topping2021UnderstandingOA} the sensitivity viewpoint on \osq{} has thus far been the predominant approach in the literature; \textbf{in this work, we seek to revisit the storage capacity viewpoint} and investigate how this issue can be avoided. We define storage capacity as the amount of information that can be stored in a node’s representation for later use: a representation is \emph{saturated} when it is unable to store any more information.

\vspace{-0.2cm}
\paragraph{Conflation With Depth}

The vast majority of existing research links \osq{} with depth. To an extent, this is justified: the bound of \Cref{eq:sens_bound} decreases exponentially with MPNN depth, and real-world graphs tend to exhibit receptive fields that grow exponentially in depth, leading to capacity quickly becoming a problem for deep MPNNs. However, alongside recent work \citep{ArnaizRodrguez2025OversmoothingH}, we highlight that \osq{} is not \emph{exclusively} a problem of depth: bottlenecks can be observed in single-layer GNNs acting on high-degree nodes -- we exploit this fact in our synthetic task of \Cref{sec:neighbor-recall-explanation}. Furthermore, in studying \osq{} only in the {\em deep regime}, much of the literature has conflated the problem with issues of {\em vanishing gradients}, themselves closely linked to the related problem of {\em over-smoothing}  \cite{Giovanni2023HowDO}. \citet{Arroyo2025OnVG} give a more precise treatment of how the issue of \osq{} relates to depth, through over-smoothing (zero collapse) and vanishing gradients. In this work we study \osq{} in the {\em shallow regime}: this allows us to isolate the issue of saturating capacity, avoiding the effects of depth on both reduced sensitivity  (\Cref{eq:sens_bound}) and vanishing gradients.

\subsection{Existing \OSq{} Tasks do not (Only) test Capacity}\label{sec:existing_osq_limitations}

\begin{wrapfigure}{r}{0.46\textwidth}
  \vspace{-0.6cm} 
  \begin{minipage}{0.46\textwidth}
    \centering
    \includegraphics[width=0.75\linewidth]{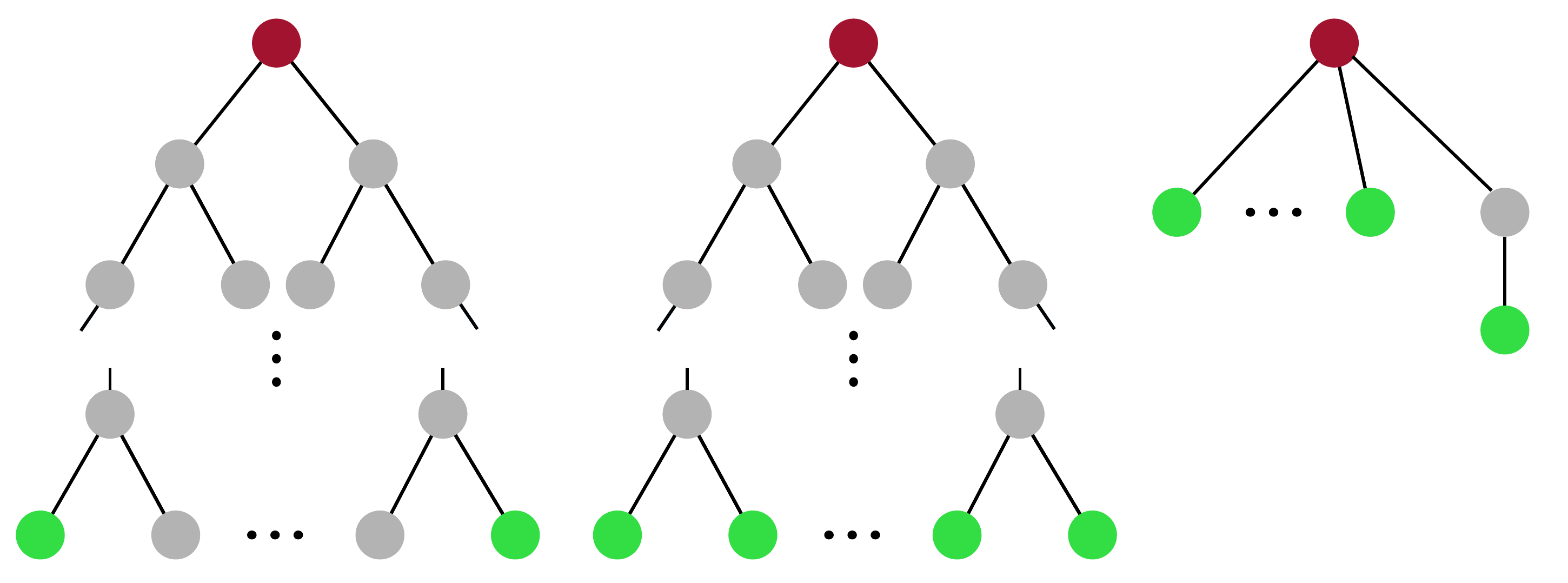}
    \caption{Computational graphs. {\bf Left}: RingTransfer \citep{Giovanni2023OnOI}.
    {\bf Middle}: \texttt{Tree-NeighborsMatch} \citep{Alon2020OnTB}.
    {\bf Right}: NAR, introduced in \Cref{sec:neighbor-recall-explanation}.
    Nodes with informative features are green, background gray. Red node is trained to solve the task.}
    \label{fig:comp_graph}
    \vspace{0.1cm}
    
    \includegraphics[width=0.9\linewidth]{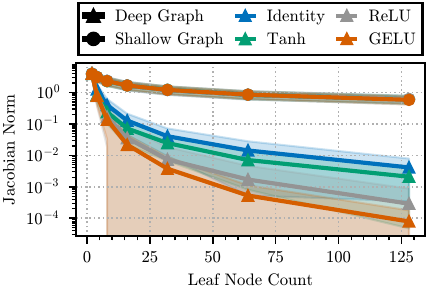}
    \caption{Log Jacobian norms. ``Deep'' graphs are binary trees of \texttt{Tree-NeighborsMatch} \citep{Alon2020OnTB}; ``Shallow'' graphs are single-level trees with the same number of leaves. A GCN of depth equal to the tree depth acts on each. Jacobian norms are $|\partial \vh_r^{(L)}/\partial \vh^{(0)}_l|_1$ for root $r$ and leaf $l$ (red/green in \Cref{fig:comp_graph}). Shaded area is standard deviation.}
    \label{fig:flat_v_deep}
  \end{minipage}
\end{wrapfigure}

An instructive way of contrasting sensitivity against capacity is via synthetic tasks. The most common of these used to assess \osq{} are the RingTransfer tasks of \citet{Giovanni2023OnOI}. The goal of these tests is for a MPNN to `transfer' features contained at a target node to a source node, across a large graph distance. Various graphs are tested, in particular a ring of nodes, but the common feature is that there exists a long shortest-path from the source to target node. All of these exhibit an exponentially growing receptive field of at least $2^k$ for $k$ layers, since each node is connected to at least two others; repeated aggregation and application of MPNN layers and nonlinearities makes this a good test of the {sensitivity}-based view of \osq{}.

However, this task is particularly ill-equipped to test the issue of storage capacity, as the \emph{only} relevant information in the graph is that of the target node, and all intermediate nodes are assigned constant vectors of ones. In this way, there is only a single node's representation worth of information to be transferred. It is unclear how much this task measures behavior found in real-world tasks: exponentially growing receptive fields will not be padded by nodes with identical representations. \Cref{fig:comp_graph} (left) visualizes the computational graph of RingTransfer, demonstrating that it is dominated by nodes containing no information. Therefore, this task exhibits a large computational bottleneck without any issues of saturating capacity: this highlights the fact that, beyond the computational bottleneck, saturated capacity is at least also dependent on the information content of the task.

\citet{Alon2020OnTB} introduced the \texttt{Tree-NeighborsMatch} task to measure capacity by propagating information from the leaf nodes of a variable-size binary tree. It shares similarities with the task we introduce in \Cref{sec:neighbor-recall-explanation} in that it controls the amount of information that is forced into a single node representation. However, it propagates this information through a deep binary tree, requiring variable-depth MPNNs. This significantly harms sensitivity: we visualize Jacobian norms of a GCN acting on a deep binary tree vs a single layer tree with matching leaf counts in \Cref{fig:flat_v_deep}, demonstrating that this sensitivity drops off far faster for deep GCNs. This is unsurprising given the bound of \Cref{eq:sens_bound}: deep GCNs must additionally contend with ``model'' squashing terms of nonlinearity and weight contraction that scale exponentially with depth. Therefore performance degradation trends are due to {both} 1) saturating capacity and 2) low sensitivity; deep tasks such as \texttt{Tree-NeighborsMatch} are impacted by both \osq{} issues, rather than isolating the issue of capacity.

\vspace{-0.2cm}

\subsection{Neighbor Associative Recall: Isolating Storage Capacity}\label{sec:neighbor-recall-explanation}

We investigate storage capacity by measuring \emph{associative recall}: this is a common approach taken in the sequence-modeling literature \citep{Ba2016UsingFW, Schlag2021LinearTA, Arora2023ZoologyMA, Jelassi2024RepeatAM}, in which the question of model storage capacity is also clearly of interest. These synthetic tasks involve presenting the model with a sequence of key value pairs followed by a query that corresponds to one of the presented keys, and the model must return the associated value.

To this end, we introduce a task that we refer to as Neighbor Associative Recall (NAR). Whereas the sequence associative recall tasks measure the ability of a model to recall previous information from a variable-length sequence, our graph adaptation is designed to measure the ability of a GNN to recall information from the previous message passing round over a variable number of neighbors.

\begin{wrapfigure}{r}{0.4\textwidth}
\vspace{-0.6cm}
  \centering
  \includegraphics[width=0.38\textwidth]{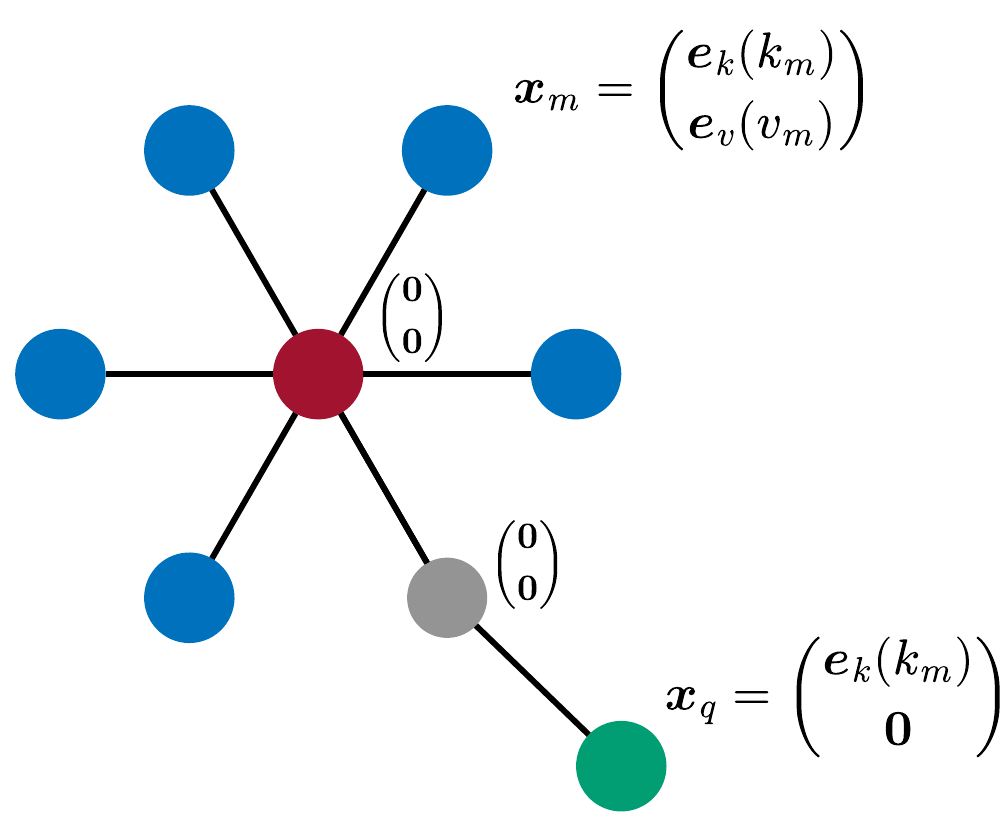}
  \caption{An example graph with $N=5$ from the NAR task. Key-value nodes are shown in blue, the central node in red and the query node in green. In this graph, $m$ is the randomly sampled index of the key-value node associated with query node $q$. The target for this graph is a one-hot vector corresponding to $v_m$.}
  \vspace{-0.25cm}
  \label{fig:capacity_example}
\end{wrapfigure}

The task is designed as follows. For a given neighborhood size $N$ we create a graph of $N +3$ nodes. This graph consists of $N$ ``neighbor'' nodes, a central node to which they are all connected, an intermediate node connected to the central node, and a ``query'' node connected only to the intermediate node. An example such graph is visualized in \Cref{fig:capacity_example}.

For a fixed neighborhood size $N$ we define a fixed set of keys and values, $N = |\sK| = |\sV|$, and a pair of learned vector embedding functions $\ve_k: \sK \to \sR^{d_{\text{emb}}}$, $\ve_v: \sV \to \sR^{d_{\text{emb}}}$ for embedding dimension $d_{\text{emb}}$. Each of the neighbor nodes $n$ has a different assigned key $k_n \in \sK$, and also a value $v_n \in \sV$, randomly sampled with replacement. The input feature vector of these nodes is a concatenation of the two learned embeddings $\vx_n = [\ve_k(k_n); \ve_v(v_n)] \in \sR^{2d_{\text{emb}}}$. The intermediate node and central node both have zero-valued feature vectors. Associated with the query node $q$ is a randomly sampled key-value node $m$; the input feature vector for the query node consists of the corresponding key embedding concatenated with a vector of zeroes, $\vx_q = [\ve_k(k_m); \boldsymbol{0}] \in \sR^{2d_{\text{emb}}}$. The model is trained such that the central node must predict the value $v_m$ associated with the sampled key node. Training is via cross-entropy loss where the target of the central node is a one-hot vector corresponding to a fixed ordering of $\sV$. This approach can be viewed as a graph adaptation of the sequence associative recall task of \citet{Schlag2021LinearTA}. Results are presented in \Cref{sec:neighbor-recall-results}. An alternative formulation of this task with a regression target is discussed in \Cref{appen:narr}.

NAR is designed such that the receptive field of the central node will comprise \emph{only} the neighbor nodes in the first layer. In order to perfectly solve the task, it must store all of the key-value information in this initial receptive field, as it is impossible to limit the scope of the information that might later be required. In the second layer, the receptive field will include the query node: now, the model must selectively recall the correct value from its immediate neighbors.

This task is novel as it assesses \osq{} in the shallow regime: MPNNs tested in \Cref{sec:neighbor-recall-results} consist of just two message passing layers. This more effectively isolates the issue of capacity, without secondary effects from low sensitivity and vanishing gradients as visualized in \Cref{fig:flat_v_deep}.
\vspace{-0.3cm}
\section{\modelname{}: Combining Graph Networks and Associative Memory}\label{sec:architecture}

Prior work on \osq{} has focused almost exclusively on mitigating sensitivity issues, often through graph rewiring \citep{gasteiger2019diffusion,gutteridge2023drew,nguyen2023revisiting}. Some work has implicitly tackled capacity \osq{} by moderating the flow of information into node representations \citep{Bresson2017ResidualGG,finkelshtein2024cooperative,errica2025adaptive} thus reducing capacity requirements, but we are unaware of any work that has attempted to \emph{increase} capacity at an architecture level. Motivated by memory-capacity gains in sequence models \citep{Ba2016UsingFW,Beck2024xLSTMEL}, we introduce associative memory into an MPNN architecture to explicitly enlarge its information-storage capacity; we measure this in the graph setting using the NAR task introduced above. We further introduce the gating scheme of \citet{Beck2024xLSTMEL} to investigate its efficacy in the graph setting, given strong sequence modeling performance. Since these adaptations are inspired in part by their successful use in xLSTM, we refer to our related graph architecture as \modelname{}.

For any node $u$ at layer $\layer{}$, in addition to the usual MPNN vector hidden state $\vh_u^{(\layer{})}$, \modelname{} maintains a matrix hidden state $\mC_u^{(\layer{})}$. The initial hidden state $\vh^{(0)}_u$ is the input node feature vector $\vx_u$. Keys and values are used to update $\mC^{(\layer{})}_u$ via an FWP-style outer product rule: these are projections of the previous vector hidden state $\vh^{(\layer{} -1)}_u$. The next vector hidden state $\vh^{(\layer{})}_u$ is determined by ``querying'' $\mC^{(\layer{})}_u$ via matrix multiplication with another projection of the previous vector hidden states.

The modified \modelname{} update equations are given below. Highlighted in blue are the differences to xLSTM. Biases correspond exactly to xLSTM (\Cref{appen:xlstm}) and are omitted for clarity.

\begin{minipage}[t]{0.48\textwidth}
\textbf{State (and normalization) updates:}
\begin{align*}
    \mC^{(\layer{})}_u &= f^{(\layer{})}_u \mC^{(\layer{} -1)}_u + \color{alvaroblue} \sum_{v \in \mathcal{N}_u^{(\layer{})} \cup \{ u \}} i^{(\layer{})}_v \vv^{(\layer{})}_v \otimes \vk^{(\layer{})}_v \\
    \vn^{(\layer{})}_u &= f^{(\layer{})}_u \vn^{(\layer{} - 1)}_u +  \color{alvaroblue} \sum_{v \in \mathcal{N}_u^{(\layer{})} \cup \{ u \}} i^{(\layer{})}_v \vk^{(\layer{})}_v \\
    m^{(\layer{})}_u &= \max{\left ( \left \{ \tilde{f}^{(\layer{})}_u + m^{(\layer{} - 1)}_u \right \} {\color{alvaroblue}\cup \left \{ \tilde{i}^{(\layer{})}_v \mid \forall v \in \mathcal{N}_u^{(\layer{})} \cup \{ u \} \right \}} \right ) }
\end{align*}
\end{minipage}
\begin{minipage}[t]{0.48\textwidth}
\textbf{Query / Key / Value computation:}
\begin{align*}
    \vq^{(\layer{})}_u &= \mW_q \color{alvaroblue}{\left [\vh^{(\layer{} - 1)}_u; \sum_{v \in \mathcal{N}_u^{(\layer{})}} \vh^{(\layer{} - 1)}_v \right ]} \\
    \vk^{(\layer{})}_u &= \frac{1}{\sqrt{d}} \mW_k \color{alvaroblue}{\vh^{(\layer{} - 1)}_u} \\
    \vv^{(\layer{})}_u &= \mW_v \color{alvaroblue}{\vh^{(\layer{} - 1)}_u}
\end{align*}
\end{minipage}

The square brackets above denote vector concatenation. Concatenating the hidden state for the node and its neighbours in this way keeps them separate and allows the query -- which will determine what is retrieved from the matrix memory -- to \emph{separately} depend on both the previous state of the node itself and the previous states of its neighbours.

\begin{minipage}[t]{0.58\textwidth}
\textbf{Gate computation:}
\begin{align*}
    i^{(\layer{})}_u &= \exp{\left( \tilde{i}^{(\layer{})}_u - m^{(\layer{})}_u \right)} 
    & \tilde{i}^{(\layer{})}_u &= \vw_i^T \color{alvaroblue}{\vh^{(\layer{} - 1)}_u} \\
    f^{(\layer{})}_u &= \exp{\left( \tilde{f}^{(\layer{})}_u + m^{(\layer{} - 1)}_u - m^{(\layer{})}_u \right)} 
    & \tilde{f}^{(\layer{})}_u &= \vw_f^T \color{alvaroblue}{\vh^{(\layer{} - 1)}_u} \\
    \vo^{(\layer{})}_u &= \sigma{\left( \tilde{\vo}^{(\layer{})}_u \right)} 
    & \tilde{\vo}^{(\layer{})}_u &= \mW_o \color{alvaroblue}{\vh^{(\layer{} - 1)}_u}
\end{align*}
\end{minipage}
\hfill
\begin{minipage}[t]{0.38\textwidth}
\textbf{Output:}
\begin{align*}
    \tilde{\vh}^{(\layer{})}_u &= \frac{\mC^{(\layer{})}_u \vq^{(\layer{})}_u}{\max{ \left \{ \left | \vn^{(\layer{}) \,\top}_u \vq^{(\layer{})}_u \right |, 1 \right \} }} \\
    \vh^{(\layer{})}_u &= \vo^{(\layer{})}_u \odot \tilde{\vh}^{(\layer{})}_u
\end{align*}
\end{minipage}

\begin{wrapfigure}{l}{0.5\textwidth}
    \vspace{-10pt}
    \centering
    \includegraphics[width=0.45\textwidth]{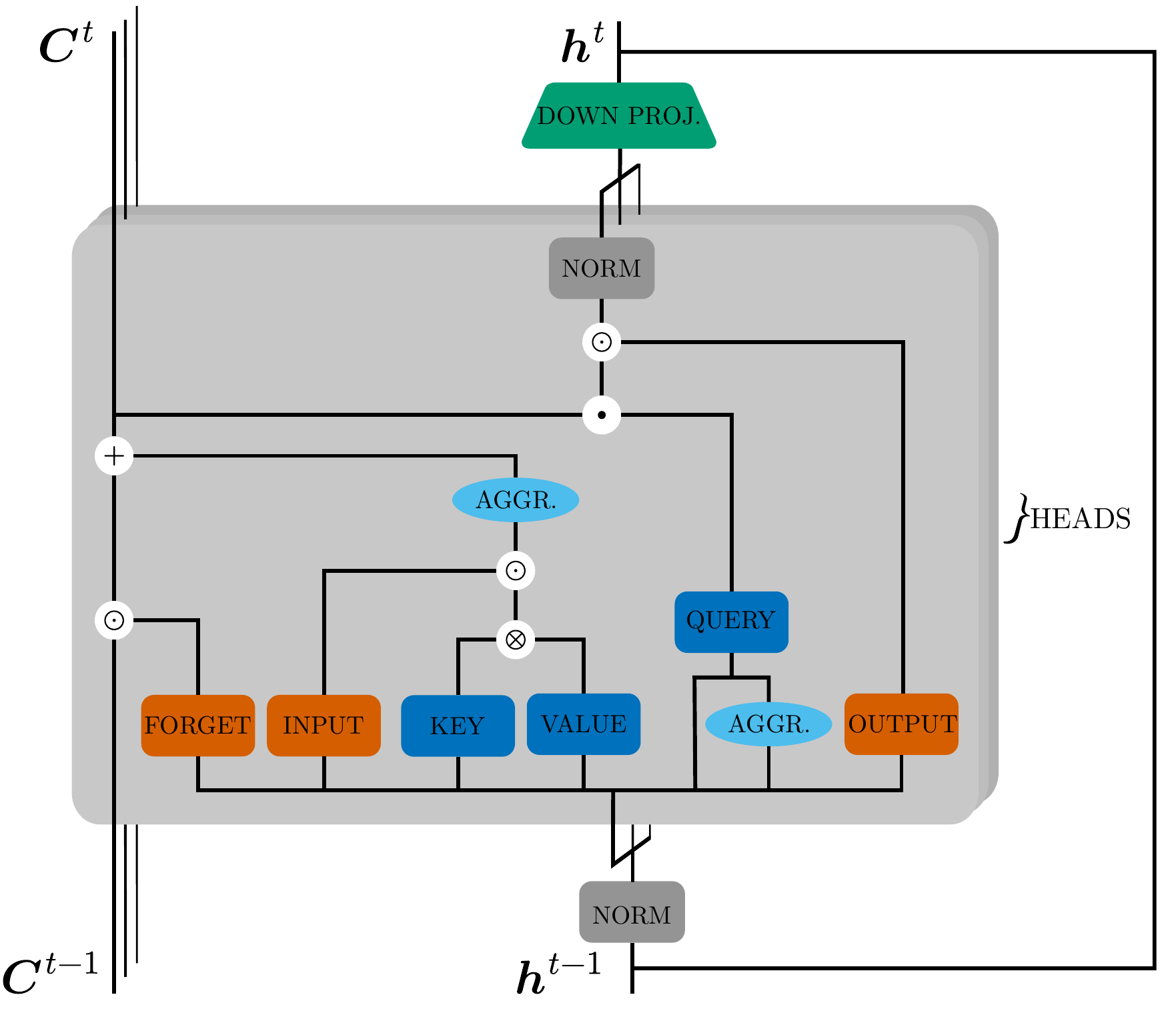}
    \caption{\modelname{} block structure. Gates shown in orange, query/key/value in dark blue. \emph{Aggr.}\@ represents aggregation across neighborhoods. Symbols $\odot, \otimes, +, \cdot$ denote Hadamard product, outer product, vector addition, matrix multiplication.}
    \label{fig:glstm_block_structure}
\end{wrapfigure}

\paragraph{Block Structure}
\citet{Arroyo2025OnVG} note that sensitivity over-squashing issues are largely caused by vanishing gradients -- a phenomenon well-explored in the sequence-modeling literature. In an attempt to address this, \modelname{} therefore uses a similar block structure to the mLSTM block upon which it is based. Of particular importance is the residual connection -- which brings the norm of the layer-wise Jacobian to the edge of chaos -- and use of input and hidden norms, which regulate the magnitude of the Jacobian norms. \Cref{fig:glstm_block_structure} visualizes the block structure of \modelname{} that we employ.

\paragraph{K-Hop Aggregation}\label{subsec:khop}

Following \citet{Arroyo2025OnVG} we combine the memory capabilities of the xLSTM block with a \emph{highly connected} message passing graph structure: employing a k-hop aggregation scheme. In this setting, each node $u$ at layer $\layer{}$ will aggregate information from the neighborhood
\[ \mathcal{N}_u^{(\layer{})} = \left \{ v \in \mathcal{V} \mid d_{\mathcal{G}}(u, v) = \layer{} \right \}, \]
where $d_{\mathcal{G}} : \mathcal{V} \times \mathcal{V} \to \mathbb{R}_{\geq 0}$ is the length of the minimal walk connecting nodes $u$ and $v$. This approach resembles that of \citet{Ding2023RecurrentDF}, but with an additional recurrence: hidden states are used as input at each step. This substantially changes the way information can propagate through the graph. Furthermore, it also has links to ChebNet \citep{defferrard2017convolutional}, which has recently been found to perform strongly on long-range tasks \citep{hariri2025return}.

This aggregation scheme appears to greatly improve \modelname{} performance: our synthetic task in \Cref{sec:neighbor-recall-results} significantly benefits from this aggregation scheme, and the ablations in \Cref{apen:ablations} demonstrate that it improves performance in all but one of the tested benchmarks. We hypothesize that -- in addition to providing a highly connected computational graph that lessens \osq{} sensitivity bottleneck issues -- this is because it also provides an extremely useful inductive bias for the \emph{recall} mechanism of \modelname{}. Information that has previously been stored in the associative memory is not then included in later message passing rounds, and later nodes are able to query this memory in isolation.

\section{Experiments}\label{sec:experiments}

\subsection{Neighbor Associative Recall}\label{sec:neighbor-recall-results}

\begin{figure}[ht]
  \centering
  \setlength{\tabcolsep}{0pt}
  \makebox[\textwidth][c]{%
  \begin{tabular}{cc}
    \subcaptionbox{Accuracy\label{fig:nar_classification_mixed_aggregation_performance}}{
      \includegraphics[height=1.1in]{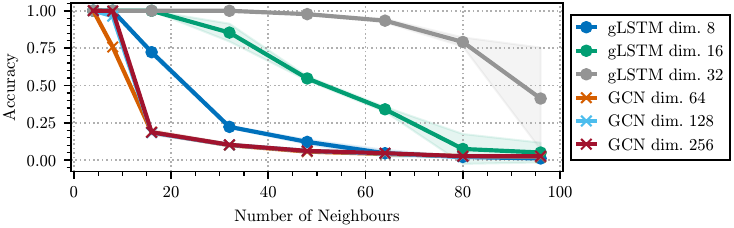}
    } &
    \subcaptionbox{Parameter counts\label{fig:trainable_params}}{
      \raisebox{0.4cm}{\includegraphics[height=0.9in]{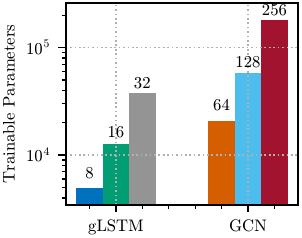}}
    }
  \end{tabular}
  }
  \caption{Test-set mean Accuracy (standard deviation shaded) for the NAR task, for \modelname{} and GCN models with various hidden dimensions shown in \Cref{fig:nar_classification_mixed_aggregation_performance}, number of trainable parameters in \Cref{fig:trainable_params}. Note that \modelname{} uses K-hop aggregation here, whereas GCN does not; see \Cref{appen:expanded_nar} for separated performance by aggregation strategy.}
  \label{fig:nar_combined}
\end{figure}

We train various models on NAR with varying neighbor count $N$, with results shown in \Cref{fig:nar_classification_mixed_aggregation_performance}. Throughout this section we compare \modelname{} using K-hop aggregation to GCN using standard aggregation, since \modelname{} performs significantly better in this task when using K-hop aggregation whereas GCN performance is harmed by K-hop. We present additional results in \Cref{appen:expanded_nar} where we separate by aggregation method and include results for a larger number of models. A comparison of the number of trainable parameters is shown in \Cref{fig:trainable_params}. Fair comparison between matrix and vector memory is nontrivial, so we select these parameter counts to ``favor'' GCN.

These results demonstrate that \modelname{} shows significantly improved recall abilities compared to GCN. \modelname{} retains perfect recall until the number of neighbors equals the memory dimension of the model: beyond this is where capacity \osq{} appears to become a problem. This agrees with intuition, since the maximum number of orthogonal key vectors (and separately, value vectors) is equal to the memory dimension. However, it is interesting to note how the performance decreases slowly as the neighbor count exceeds this limit, particularly for higher memory dimensions: this appears to be a graph analog of the ``graceful saturation'' described by \citet{Smolensky1990TensorPV}. By contrast, capacity \osq{} starts much earlier at just $N=8$ for the largest GCN model tested.

\subsection{How does Capacity Relate to Sensitivity?}\label{sec:nar_sensitivity}

In this section, we investigate empirically how capacity \osq{} -- as measured by performance on NAR -- relates to sensitivity \osq{}.

\begin{figure}[ht]
  \centering
  \setlength{\tabcolsep}{0pt}
  \makebox[\textwidth][c]{%
  \begin{tabular}{cc}
    \subcaptionbox{\hspace{0mm}Jacobian norms\hspace{-5mm}\label{fig:nar_jacobian_norms}}{
      \includegraphics[height=1.1in]{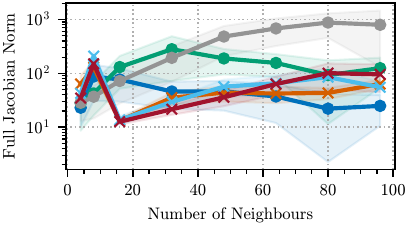}
    } &
    \subcaptionbox{\hspace{0mm}Jacobian norm ratios\hspace{15mm}\label{fig:nar_jacobian_norm_ratios}}{
      \includegraphics[height=1.1in]{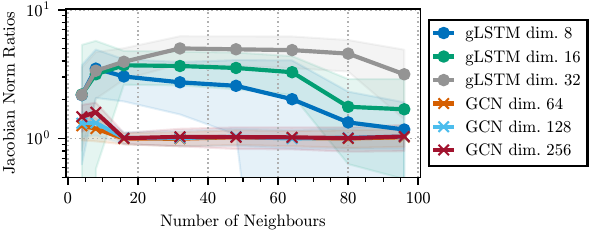}
    }
  \end{tabular}
  }
  \caption{{\bf Left}: Mean Jacobian norms for different \modelname{} and GCN models, with varying number of neighbors in the NAR task. {\bf Right}: Mean ratio between the Jacobian norms of the selected (key corresponds to query) to background (key is different from query) neighbor nodes, for varying model dimensions. Standard deviation shaded in both plots.}
  \label{fig:nar_jacobians_combined}
\end{figure}

We directly measure the Jacobian norm of \citet{Topping2021UnderstandingOA,Giovanni2023OnOI}, computing the sensitivity of the output feature vector on the central (output) node $c$ to the input vectors on the key-value neighbor nodes $n$, $|\partial \vh_{c}^{(2)} / \partial \vx_{n} |_1$. These results are visualized in \Cref{fig:nar_jacobian_norms}.

We see therefore that sensitivity, as measured by the Jacobian norm, does not correlate with NAR performance. Given that NAR performance degradation is due to capacity \osq{}, we therefore observe that \textbf{capacity \osq{} can occur without sensitivity \osq{}}. This is clear from the fact that 1) sensitivity increases consistently for GCN models above $N=16$ to the point where it matches initial sensitivity, despite no increase in performance and 2) sensitivity for \modelname{} tends to carry on increasing beyond where performance starts to degrade. We note these trends -- as with all observations we make in this section -- hold true for the NAR regression task in \Cref{appen:narr_sensitivity}.

However, if we examine the difference in Jacobian norms between the neighbor nodes which are \emph{selected} (those which have a key corresponding to the query node) vs \emph{background}, we see trends that align with our notion of capacity. \Cref{fig:nar_jacobian_norms} visualizes the ratio of Jacobian norms for selected nodes to that for background nodes. We observe that for all GCN models this ratio quickly falls to unity at the point where capacity \osq{} starts to occur, and \modelname{} ratios consistently plateau -- and start to slowly decrease -- at their memory dimension, similarly coinciding with capacity \osq{}. It appears therefore that capacity \osq{} harms a model's ability to be selectively sensitive to different nodes in the NAR task.

Another \osq{} sensitivity metric is that of \citet{Giovanni2023HowDO}, who introduce the \emph{maximal mixing} metric. For node-level function $\boldsymbol{Y}: \mathbb{R}^{n \times d} \to \mathbb{R}^{n \times d}$, the mixing of features associated with nodes $u,v$ at a given node $i$ is defined as
$$\mathop{\text{mix}}_{\boldsymbol{Y}} (i, v, u) = \max_{\boldsymbol{X}} \left \lVert \frac{\partial^2 \left ( \boldsymbol{Y} \left ( \boldsymbol{X} \right ) \right )_i}{\partial \vx_u \partial \vx_v} \right \rVert.$$

Although motivated through intuition of mixing, we observe the mixed partial derivative can equally be viewed as a composition of partial derivatives quantifying \emph{selective sensitivity} - how much the sensitivity with respect to one node feature varies with respect to another node feature. In this respect, we expect it to be highly relevant to the sensitivity ratios visible in \Cref{fig:nar_jacobians_separated,fig:nar_jacobian_norm_ratios}.

\begin{figure}[ht]
    \centering
    \includegraphics[height=1.2in]{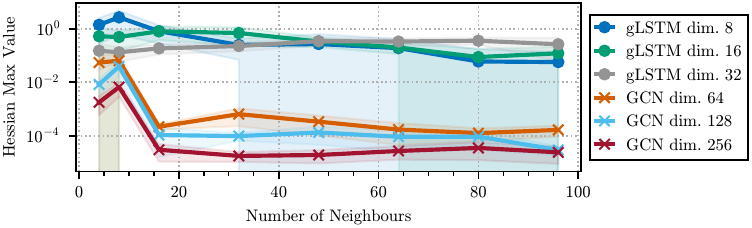}
    \caption{Mean of the maximum Hessian values for different \modelname{} and GCN models, averaged across test set examples and different neighbor nodes. Standard deviation shaded.}
    \label{fig:nar_hessians_combined}
\end{figure}

To study this empirically for NAR, we take the maximum over the measured Hessians for different models. These Hessian 3-tensors are large, so we further limit to a subset of the overall tensor in order to compute them on available hardware: we are most interested in how the output sensitivity to the neighbor \emph{value} vectors varies with the \emph{query} vector, so we limit to the corresponding input dimensions. For the central, neighbor and query nodes $c,n,q$ this adapted mixing metric is
$$\mathop{\text{mix}} (c, n, q) = \max_{\substack{0 \leq \alpha < N, \\ 0 \leq \beta < d_{\text{emb}}, \\ d_{\text{emb}} \leq \gamma < 2d_{\text{emb}}}} \left \lvert \frac{\partial^2 \left (\vh^{(2)}_c \right )_{\alpha}}{\partial \left (\vx_q \right )_{\beta} \partial \left (\vx_n \right )_{\gamma}} \right \rvert,$$
which we plot in \Cref{fig:nar_hessians_combined}. We see that \modelname{} consistently exhibits greater maximum Hessian values than GCN, and that this collapses for GCN models above 8 neighbors, consistent with the drop in performance. As with the Jacobian ratios, we see plateauing and slow decrease of maximum Hessian values above the memory dimension, but these trends are less pronounced.

\newpage
\subsection{Long Range Benchmarks}

\begin{table*}[ht]
    \centering
    \caption{Mean and standard deviation of $\log_{10}(\text{MSE})$, averaged over 4 random weight initializations on the GPP tasks from \citet{Gravina2022AntiSymmetricDA}, from which we report baselines. See \Cref{apen:extended_gpp} for discussion of baseline choice. \textbf{Top} score in bold, \underline{second} underlined. Lower is better.}
    \label{tab:gpp_results}
    \begin{tabular}{lccc}
        \toprule
        \textbf{Method} & \textbf{Diam.} & \textbf{Ecc.} & \textbf{SSSP} \\
        \midrule
        GCN        & 0.742 ± 0.047 & 0.846 ± 0.003 & 0.950 ± 0.000 \\
        GAT        & 0.822 ± 0.075 & 0.791 ± 0.022 & 0.695 ± 0.150 \\
        GraphSAGE & 0.865 ± 0.40 & 0.286 ± 0.184 &  0.786 ± 0.021 \\
        GIN & 0.613 ± 0.099 & 0.950 ± 0.001  & -0.541 ± 0.419 \\
        GCNII & 0.529 ± 0.057 & 0.764 ± 0.036  & -1.132 ± 0.013 \\
        DGC & 0.603 ± 0.005 & 0.826 ± 0.003 & -0.148 ± 0.023 \\
        GRAND & 0.672 ± 0.049 & 0.660 ± 0.139 & -0.094 ± 0.340 \\
        A-DGN & \underline{-0.546 ± 0.033} & \underline{0.305 ± 0.118} & {\bf -3.402 ± 0.137} \\
        \midrule
        \modelname{} (ours) & {\bf -0.715 ± 0.030} & {\bf -4.036 ± 0.311} & -2.836 ± 0.178 \\
        - K-hop & 0.042 ± 0.123 & 0.673 ± 0.021 & \underline{-3.377 ± 0.142}  \\
        \bottomrule
    \end{tabular}
\end{table*}

\begin{wraptable}{r}{0.5\textwidth} 
    \centering
    \vspace{-0.45cm}
    \caption{Mean and standard deviation on LRGB \citep{Dwivedi2022LongRG}, averaged over four random weight initializations. Baselines from the LRGB reevaluation of \citet{Tonshoff2023WhereDT} and K-hop methods from \citet{Arroyo2025OnVG}. All methods adhere to a 500k parameter limit. \textbf{Top} score in bold, \underline{second} underlined.}
    \label{tab:lrgb_results}
    \addtolength{\tabcolsep}{-0.4em}
    \begin{tabular}{lcc}
        \toprule
        \textbf{Method} & \textbf{Peptides-Func} & \textbf{Peptides-Struct} \\
        \cmidrule(lr){2-3}
        & \textbf{AP ($\uparrow$)} & \textbf{MAE ($\downarrow$)} \\
        \midrule
        GCN & 0.6860±0.0050 & {\bf 0.2460±0.0007}  \\
        GatedGCN & 0.6765±0.0047 & 0.2477±0.0009 \\
        GINE & 0.6621±0.0067 & \underline{0.2473±0.0017} \\
        GPS & 0.6534±0.0091 & 0.2509±0.0014 \\
        \midrule
        \textit{K-hop methods} \\
        kGCN-SSM & \underline{0.6902±0.0022} & 0.2581±0.0003 \\
        DRew-GCN & {0.6804±0.0144} & 0.2766±0.0019 \\
        \midrule
        gLSTM (ours) & {\bf 0.7250±0.0023} & 0.2527±0.0015 \\
        \bottomrule
    \end{tabular}
\end{wraptable}
We evaluate \modelname{} on the Graph Property Prediction (GPP) tasks from \citet{Gravina2022AntiSymmetricDA} and the Long Range Graph Benchmark (LRGB) from \citet{Dwivedi2022LongRG}. These benchmarks are both designed to require long range interactions to solve, and thus are an interesting test of the ability of \modelname{} to overcome \osq{} and over-smoothing in real world tasks in order to facilitate long range interactions. Performance is reported in \Cref{tab:gpp_results} and \Cref{tab:lrgb_results} respectively.

\modelname{} achieves comfortably state of the art results on the Diameter and Eccentricity GPP tasks, and very strong performance on SSSP; notably SSSP is the only tested task in which k-hop decreases performance. LRGB results show that \modelname{} achieves strong performance in Peptides-Func but relatively weak performance on Peptides-Struct. We hypothesize that the weaker performance on Peptides-Struct may be due to long-range interactions being less relevant for this task, which is very effectively solved by a few-layer GCN. See \Cref{apen:ablations} for \modelname{} ablations on these benchmarks and \Cref{apen:hyperparams} for details around hyperparameters used.

\vspace{-0.1cm}
\section{Conclusion}

In this work, we revisit \osq{}, disambiguating two bottleneck-related issues of sensitivity \osq{} and capacity \osq{}. We introduce a synthetic task that measures capacity \osq{} in isolation and we show that associative memory can improve MPNN capacity. The resulting architecture achieves strong results on real-world benchmarks.

\paragraph{Future Work }Many avenues remain open. Whereas the sensitivity issue of \osq{} has a mathematical basis via the node Jacobian, to our knowledge, the capacity issue does not. Theoretically quantifying this capacity could afford similar directions to those explored via sensitivity, establishing links to topology and model properties. With regards to architecture, we translate to a graph setting the gating and associative memory of xLSTM but do not retain the efficiency and parallel training, leaving open future work on more efficient MPNNs: we highlight that the recent work of \citet{poppel2025plstm} achieves these efficiency gains in the specific case of directed acyclic graphs. Another potential avenue would be to apply our findings to prevention of issues of over-mixing and representational collapse  \citep{barbero2024transformers, barbero2025llms} in Transformer architectures.

\section*{Reproducibility Statement}

We make available all of our code and experiment configurations to aid reproduction of results. Our experiments utilize the widely-used PyTorch Geometric GraphGym \citep{you2020design} framework which defines a standard framework for MPNN research.

For easiest reproduction of our results, please consult the \verb|readme| in the code repository provided in \Cref{apen:additional_experiments}. The repository includes all necessary information to run the experiments: in particular, configs containing the hyperparameters used (also reported in \Cref{apen:hyperparams}) and code for all plots used in the paper.


\ificlrfinal
    \subsubsection*{Acknowledgments}
    HB acknowledges funding support from the EPSRC Centre for Doctoral Training in Autonomous Intelligent Machines and Systems No. EP/S024050/1. MB is partially supported by the EPSRC Turing AI World-Leading Research Fellowship No. EP/X040062/1 and EPSRC AI Hub No. EP/Y028872/1. XD acknowledges support from the Oxford-Man Institute of Quantitative Finance and EPSRC No. EP/T023333/1.
\else
    \ifarxivfinal
        \subsubsection*{Acknowledgments}
        HB acknowledges funding support from the EPSRC Centre for Doctoral Training in Autonomous Intelligent Machines and Systems No. EP/S024050/1. MB is partially supported by the EPSRC Turing AI World-Leading Research Fellowship No. EP/X040062/1 and EPSRC AI Hub No. EP/Y028872/1. XD acknowledges support from the Oxford-Man Institute of Quantitative Finance and EPSRC No. EP/T023333/1.
    \else
    \fi
\fi

\bibliography{iclr2026_conference}
\bibliographystyle{iclr2026_conference}

\newpage
\appendix
\section{Relationship to Computational (and Topological) Bottlenecks}\label{appen:only_computational_tree}
\citet{ArnaizRodrguez2025OversmoothingH} identify a separation in the \osq{} literature between the initial work of \citet{Alon2020OnTB} and the later work of \citet{Topping2021UnderstandingOA,Giovanni2023OnOI}. They point out that the original \osq{} definition of \citep{Alon2020OnTB} was associated with the computational tree, and \citet{Topping2021UnderstandingOA} later connected \osq{} to the existence of topological bottlenecks. They suggest that these definitions constitute different problems, and that -- as a community -- we should discard the term ``\osq{}'' and separate it into (at least) two separate terms: (1) computational tree bottlenecks and (2) topological bottlenecks (in the underlying graph).

This makes a valuable point: topological bottlenecks may or may not be involved when structural issues exist with the computational tree. The authors fairly point out that the umbrella term ``\osq{}'' has sometimes hidden some of the complexity of the problem, and the field may benefit from work identifying explicitly whether it is dealing with topological bottlenecks (as is the case with e.g. rewiring) or bottlenecks of the computational tree (e.g. adaptive message passing). We're unsure that framing these as \emph{separate} is helpful, since topological bottlenecks must be mediated through the computational tree in order to impact message passing, but the point stands that some methods to combat \osq{} specifically target topological bottlenecks (and as such any changes to the computational graph are implicit), and some methods target general computational bottlenecks, and that this is not always clear.

Despite the complexity, we believe the term ``\osq{}'' still has utility as an umbrella term that describes issues that arise from the presence of bottlenecks and depth in the computational tree. The distinction made by \citet{ArnaizRodrguez2025OversmoothingH} clarifies that sometimes these issues arise due to topological bottlenecks in the underlying graph, and sometimes they do not. The relevance of our work is then in exploring \emph{how} this structure manifests as performance issues, and the main body of our paper argues that this is due to separate issues of \textbf{capacity} and \textbf{sensitivity}.

Precisely \emph{how} issues of capacity and sensitivity relate to the structure of the computational tree and the presence of bottlenecks is complex, although we take initial steps to clarify this in our paper. It would \textbf{not} be correct to suggest for example that capacity issues correspond to computational bottlenecks and low sensitivity to topological bottlenecks. In particular:
\begin{itemize}
    \item Our RingTransfer discussion of \Cref{sec:existing_osq_limitations} highlights that computational bottlenecks can exist without capacity issues.
    \item \citet{ArnaizRodrguez2025OversmoothingH} highlight that low sensitivity does not necessarily imply the existence of computational tree bottlenecks (or topological bottlenecks).
    \item On the other hand, both topological bottlenecks and general computational bottlenecks can cause low sensitivity. This is clear since the \Cref{eq:sens_bound} defines an upper bound on sensitivity in terms of the computational tree (defined by successive powers of the message-passing matrix), and we refer the reader to the work of \citet{Topping2021UnderstandingOA} for proofs in the case of specifically topological bottlenecks.
\end{itemize}

Additionally, while the discussion in this Appendix has so far focused on the computational tree, we highlight that the causes of the failure modes discussed in our work extend beyond structural issues: issues of capacity also depend at least on (1) information content of the task and (2) storage capacity of the model, as explored in  \Cref{sec:existing_osq_limitations,sec:neighbor-recall-explanation} respectively, and issues of sensitivity are highly dependent on the model dynamics, as explored in e.g. \citet{Arroyo2025OnVG,heilig2025porthamiltonian,Gravina2022AntiSymmetricDA,gravina2025oversquashing}.
\section{xLSTM Update Equations}\label{appen:xlstm}
\citet{Beck2024xLSTMEL} initially designed xLSTM as a combination of sLSTM blocks and mLSTM blocks - using scalar memory and matrix (associative) memory respectively. However, their follow up work \citep{Beck2025xLSTM7A} uses only mLSTM blocks, and these form the inspiration for \modelname{}. Therefore, we will exclusively introduce the mLSTM block update equations in this section.

The update equations, presented in a similar manner to \Cref{sec:architecture}, are given below.

\textbf{State (and normalization) updates:}
\begin{align}
    \mC^t &= f^t \mC^{t-1} + i^t\vv^t \otimes \vk^{t} \\
    \vn^t &= f^t \vn^{t-1} + i^t \vk^t \\
    m^t &= \max{\left ( \tilde{f}^t + m^{t-1}, \tilde{i}^t \right )}
\end{align}

\textbf{Query / Key / Value computation:}
\begin{align}
    \vq^t &= \mW_q \vx^t + \vb_q \\
    \vk^t &= \frac{1}{\sqrt{d}} \mW_k \vx^t + \vb_k \\
    \vv^t &= \mW_v \vx^t + \vb_v
\end{align}

\textbf{Gate computation:}
\begin{align}
    i^t &= \exp{\left( \tilde{i}^t - m^t \right)} 
    & \tilde{i}^t &= \vw_i^T \vx^t + b_i \\
    f^t &= \exp{\left( \tilde{f}^t + m^{t-1} - m^t \right)} 
    & \tilde{f}^t &= \vw_f^T \vx^t + b_f \\
    \vo^t &= \sigma{\left( \tilde{\vo}^t \right)} 
    & \tilde{\vo}^t &= \mW_o \vx^t + \vb_o
\end{align}

\textbf{Output:}
\begin{align}
    \tilde{\vh}^t &= \mC^t \vq^t / \max{ \left \{ \left | \vn^{t\,T} \vq^t \right |, 1 \right \} } \\
    \vh^t &= \vo^t \odot \tilde{\vh}^t
\end{align}
\section{Additional Experiments}\label{apen:additional_experiments}
Our code for reproducing all experimental results in the main paper and appendices is publicly available at 
\ificlrfinal
    \url{https://github.com/HughBlayney/gLSTM}
\else
    \ifarxivfinal
        \url{https://github.com/HughBlayney/gLSTM}
    \else
        \url{https://anonymous.4open.science/r/GNN-xLSTM-9A2D}
    \fi
\fi
.

\subsection{GPP Baselines}\label{apen:extended_gpp}

We note that due to a subtle PyTorch issue in the original GPP code implementation, normalization is not applied to the dataset targets. A refactor appears to have unknowingly fixed this issue in later iterations of the code so later experiments are run on a normalized variant of the dataset. Unfortunately, this results in unfair comparison, as results can be substantially different between the two variants of the dataset.

Therefore, we test only on the baselines provided in the original GPP paper \citep{Gravina2022AntiSymmetricDA}, as we are confident these use the un-normalized variant of the dataset, and this provides us with the largest number of baselines to test against. We additionally ensure that our method uses the same, un-normalized GPP variant.

\subsection{Ablations}\label{apen:ablations}
To identify what elements of the \modelname{} architecture are most important for performance on these benchmarks, we perform ablations on the GPP and LRGB datasets. For LRGB, a task with a parameter limit, we ablate in two different settings: the first is simply removing the ablated component, for which results are presented in \Cref{tab:lrgb_ablations_fixed_dim}. The second is to scale the hidden dimension $\vh$ to keep the parameter count as close as possible to the 500k limit - i.e.\@ when removing gating, this will correspondingly increase the hidden dimension. We include these experiments as they more accurately represent the reality of testing a model variant on a task with a parameter limit; these results are presented in \Cref{tab:lrgb_ablations_variable_dim}. These two ablation settings show very similar results. GPP ablations are presented in \Cref{tab:gpp_ablations}.

We see that ablating gating only significantly reduces performance on Peptides-Func - other than this, it either leaves performance the same or in some cases, improves performance (GPP Ecc.\@ in particular).

\begin{table*}[htbp]
    \centering
    \caption{Ablation of gLSTM performance on Diam, Ecc, and SSSP from the GPP benchmark. Mean and standard deviation are reported, averaged over four random weight initializations. Other than ablation, all other model settings are held constant; thus ablations with gating removed have reduced parameter count.}
    \label{tab:gpp_ablations}
    \begin{tabular}{lccc}
        \toprule
        \textbf{Model} & \textbf{Diam.} & \textbf{Ecc.} & \textbf{SSSP} \\
        \midrule
        \modelname{} & -0.715 $\pm$ 0.030 & -4.036 $\pm$ 0.311 & -2.836 $\pm$ 0.178  \\
        \midrule
        - Output gate & -0.70 $\pm$ 0.05 & -3.71 $\pm$ 0.16 & -2.77 $\pm$ 0.19  \\
        - Input gate & -0.75 $\pm$ 0.01 & -4.72 $\pm$ 0.36 & -3.27 $\pm$ 0.16  \\
        - Forget gate & -0.71 $\pm$ 0.03 & -4.30 $\pm$ 0.21 & -3.14 $\pm$ 0.07  \\
        - All gates & -0.75 $\pm$ 0.03 & -4.14 $\pm$ 0.42 & -3.16 $\pm$ 0.15  \\
        - K-hop aggregation & 0.04 $\pm$ 0.12 & 0.67 $\pm$ 0.02 & -3.38 $\pm$ 0.14  \\
        \bottomrule
    \end{tabular}
\end{table*}

\begin{table*}[htbp]
    \centering
    \caption{Ablation of gLSTM performance on Peptides-Func and Peptides-Struct from the LRGB. Mean and standard deviation are reported, averaged over four random weight initializations. Other than ablation, all other model settings are held constant; thus ablations with gating removed have reduced parameter count.}
    \label{tab:lrgb_ablations_fixed_dim}
    \begin{tabular}{lcc}
        \toprule
        \textbf{Model} & \textbf{Peptides-Func} & \textbf{Peptides-Struct} \\
        \cmidrule(lr){2-3}
        & \textbf{AP ($\uparrow$)} & \textbf{MAE ($\downarrow$)} \\
        \midrule
        gLSTM & 0.7250 $\pm$ 0.0023 & 0.2527 $\pm$ 0.0015  \\
        \midrule
        - Output gate & 0.7086 $\pm$ 0.0049 & 0.2540 $\pm$ 0.0016  \\
        - Input gate & 0.7186 $\pm$ 0.0029 & 0.2524 $\pm$ 0.0027  \\
        - Forget gate & 0.7236 $\pm$ 0.0063 & 0.2522 $\pm$ 0.0011  \\
        - All gates & 0.7180 $\pm$ 0.0088 & 0.2526 $\pm$ 0.0012  \\
        - Positional encoding & 0.7208 $\pm$ 0.0072 & 0.2539 $\pm$ 0.0036  \\
        - K-hop aggregation & 0.6030 $\pm$ 0.0096 & 0.2638 $\pm$ 0.0010  \\
        \bottomrule
    \end{tabular}
\end{table*}
\begin{table*}[htbp]
    \centering
    \caption{Ablation of gLSTM performance on Peptides-Func and Peptides-Struct from the LRGB. Mean and standard deviation are reported, averaged over four random weight initializations. All methods adhere to a 500k parameter limit such that hidden dimension varies to keep parameter count as close to this as possible.}
    \label{tab:lrgb_ablations_variable_dim}
    \begin{tabular}{lcc}
        \toprule
        \textbf{Model} & \textbf{Peptides-Func} & \textbf{Peptides-Struct} \\
        \cmidrule(lr){2-3}
        & \textbf{AP ($\uparrow$)} & \textbf{MAE ($\downarrow$)} \\
        \midrule
        gLSTM & 0.7250 $\pm$ 0.0023 & 0.2527 $\pm$ 0.0015  \\
        \midrule
        - Output gate & 0.7202 $\pm$ 0.0056 & 0.2537 $\pm$ 0.0011  \\
        - Input gate & 0.7193 $\pm$ 0.0110 & 0.2518 $\pm$ 0.0027  \\
        - Forget gate & 0.7148 $\pm$ 0.0107 & 0.2545 $\pm$ 0.0043  \\
        - All gates & 0.7188 $\pm$ 0.0060 & 0.2528 $\pm$ 0.0035  \\
        - Positional encoding & 0.7211 $\pm$ 0.0062 & 0.2601 $\pm$ 0.0017  \\
        - K-hop aggregation & 0.6030 $\pm$ 0.0096 & 0.2638 $\pm$ 0.0010  \\
        \bottomrule
    \end{tabular}
\end{table*}

\subsection{Hyperparameters}\label{apen:hyperparams}

In \Cref{tab:gpp_hyperparams,tab:lrgb_hyperparams} we present the hyperparameter sweeps and chosen hyperparameters for GPP and LRGB respectively.

\begin{table*}[htbp]
    \centering
    \caption{Hyperparameter sweeps for \modelname{} on LRGB tasks. In bold are the hyperparameters that achieved the best validation set performance, and thus were those used in the main results of the paper. Note that hidden dimension was not directly swept over, as this was maximized for each configuration such that the model remained within the 500k parameter budget. Due to compute limitations, hyperparameter sweeps were not exhaustive, but used \emph{Weights and Biases} Bayesian Optimization routine with Hyperband early termination.}
    \label{tab:lrgb_hyperparams}
    \begin{tabular}{lcc}
        \toprule
        \textbf{Hyperparameter} & \textbf{Peptides-Func} & \textbf{Peptides-Struct} \\
        \midrule
        Memory Dimension & 8, 16, {\bf 32} & 8, {\bf 16}, 32  \\
        Number of Heads & 1-{\bf 2}-8 & 1-{\bf 5}-8  \\
        Message Passing Layers & 10-{\bf 27}-50 & 4-{\bf 23}-40  \\
        Input Norm Type & {\bf Layer} & {\bf Layer}, None \\
        Hidden Norm Type & {\bf Group} & {\bf Group} \\
        Act.\@ Func.\@ (between block) & GeLU, ReLU, {\bf None} & GeLU, {\bf ReLU}, None  \\
        Dropout & {\bf 0.1}  & {\bf 0.0}, 0.1, 0.2  \\
        Hidden Dimension & {\bf 45} & {\bf 42} \\
        \bottomrule
    \end{tabular}
\end{table*}

\begin{table*}[htbp]
    \centering
    \caption{Hyperparameter sweeps for \modelname{} on GPP tasks. In bold are the hyperparameters that achieved the best validation set performance, and thus were those used in the main results of the paper. Hyperparameters were tested exhaustively via grid search.}
    \label{tab:gpp_hyperparams}
    \begin{tabular}{lccc}
        \toprule
        \textbf{Hyperparameter} & \textbf{Diam.} & \textbf{Ecc.} & \textbf{SSSP} \\
        \midrule
        Memory Dimension & 8, {\bf 16} & {\bf 8}, 16 & 8, {\bf 16}  \\
        Number of Heads & {\bf 1}, 2, 3, 4 & 1, 2, 3, {\bf 4} & {\bf 1}, 2, 3, 4  \\
        Message Passing Layers & 1, 5, 10, {\bf 20} & 1, 5, {\bf 10}, 20 & 1, 5, {\bf 10}, 20  \\
        Input Norm Type & {\bf None} & {\bf None} & {\bf None} \\
        Hidden Norm Type & {\bf Group} & {\bf Group} & {\bf Group} \\
        Act.\@ Func.\@ (between block) & Tanh, {\bf ReLU}, None & Tanh, {\bf ReLU}, None & {\bf Tanh}, ReLU, None  \\
        Dropout & {\bf 0.0}  & {\bf 0.0} & {\bf 0.0}  \\
        Hidden Dimension & {\bf 10}, 20, 30 & 10, {\bf 20}, 30 & {\bf 10}, 20, 30 \\
        \bottomrule
    \end{tabular}
\end{table*}

\subsection{Oversmoothing and Long Range Dependencies}

\begin{figure}[h]
\begin{center}
\includegraphics[width=0.6\linewidth]{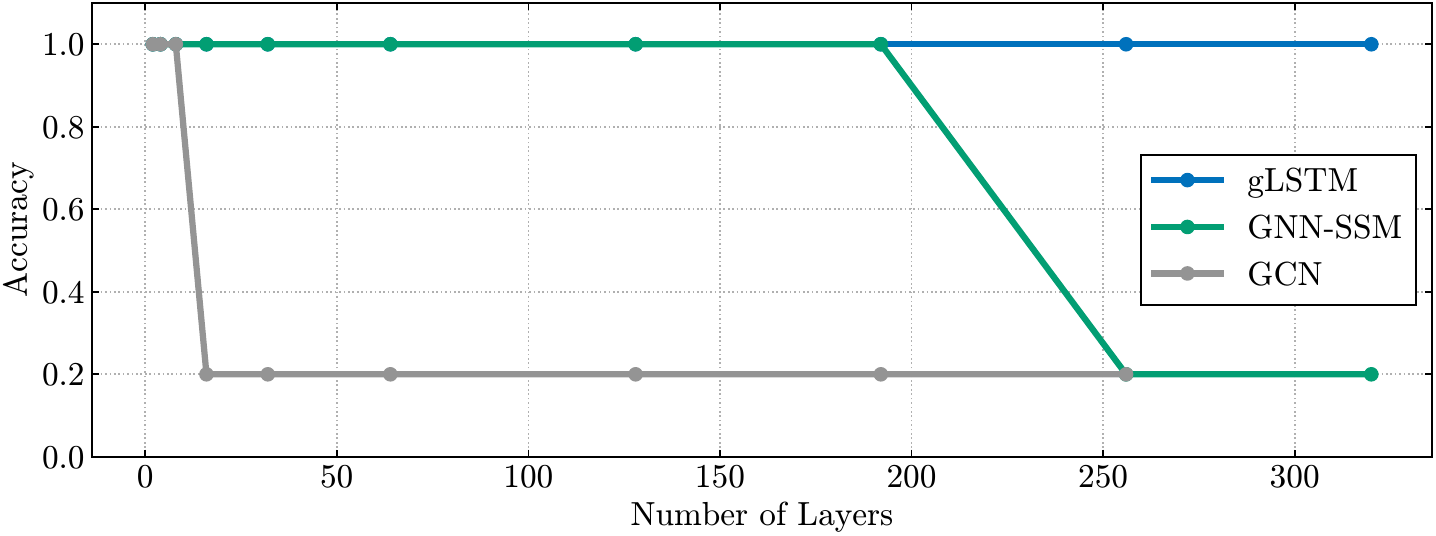}
\end{center}
\caption{Performance on the RingTransfer task.}\label{fig:ring_transfer}
\end{figure}

We test empirically that \modelname{} is able to learn long range dependencies by evaluating on the RingTransfer task introduced in \citet{Giovanni2023OnOI}. Results for \modelname{}, GCN and GNN-SSM \citep{Arroyo2025OnVG} are shown for various ring sizes (and corresponding number of message passing layers) in \Cref{fig:ring_transfer}.

\subsection{Additional NAR Classification Results}\label{appen:expanded_nar}

In this section, we present additional results from the NAR task presented in the main body of the paper.

\begin{figure}[h]
    \centering
    \includegraphics[width=0.8\linewidth]{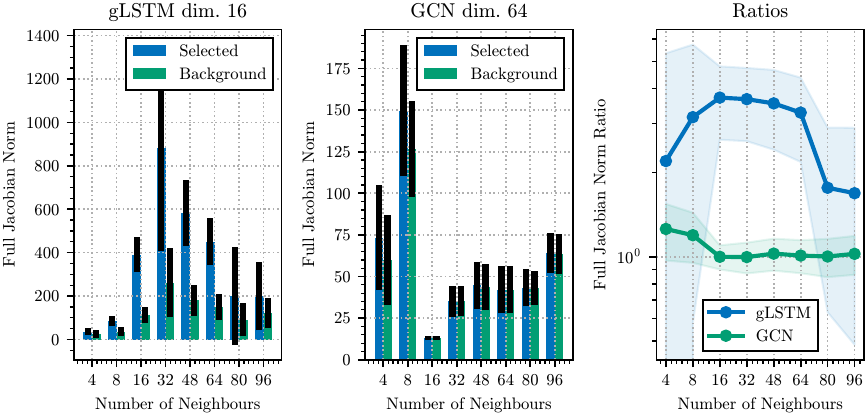}
    \caption{{\bf Left}: Mean Jacobian norms for \modelname{} of memory dimension 16, with varying number of neighbors in the NAR task, separated by whether the neighbor node corresponds to the given query (selected) or not (background). {\bf Middle}: Same, with GCN of hidden dimension 64. {\bf Right}: Mean ratios of Jacobian norms for selected nodes to background nodes, for these two models. Standard deviation visualized in bar chart error bars and line chart shaded area.}
    \label{fig:nar_jacobians_separated}
\end{figure}

We first visualize the Jacobian norms - separated by selected vs background nodes - for the mixed aggregation strategies used in the main paper in \Cref{fig:nar_jacobians_separated}. This is, in effect, the more granular plot of \Cref{fig:nar_jacobian_norm_ratios}.

\begin{figure}
    \centering
    \includegraphics[width=0.6\linewidth]{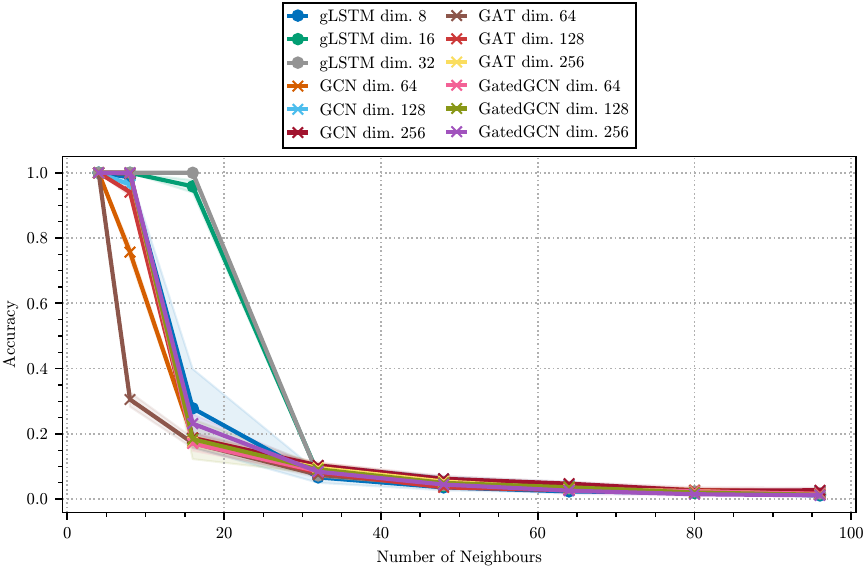}
    \caption{NAR Accuracy where all models do \emph{not} use K-hop aggregation, for an expanded set of models.}
    \label{fig:nar_classification_expanded_performance_no_k_hop}
\end{figure}

\begin{figure}
    \centering
    \includegraphics[width=0.6\linewidth]{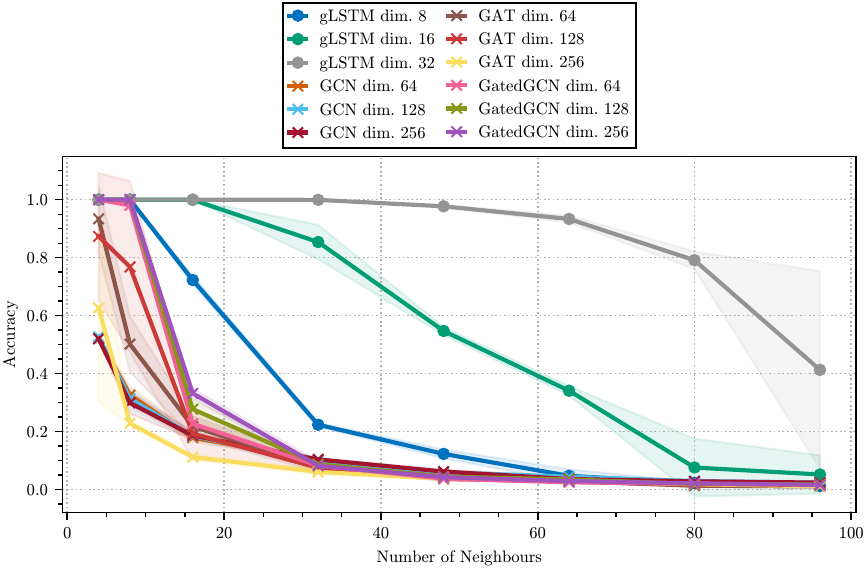}
    \caption{NAR Accuracy where all models \emph{do} use K-hop aggregation, for an expanded set of models.}
    \label{fig:nar_classification_expanded_performance_k_hop}
\end{figure}

We next separate out no-K-hop and K-hop aggregation, and plot results for a larger set of models, in \Cref{fig:nar_classification_expanded_performance_no_k_hop,fig:nar_classification_expanded_performance_k_hop} respectively.

\begin{figure}
    \centering
    \includegraphics[width=0.5\linewidth]{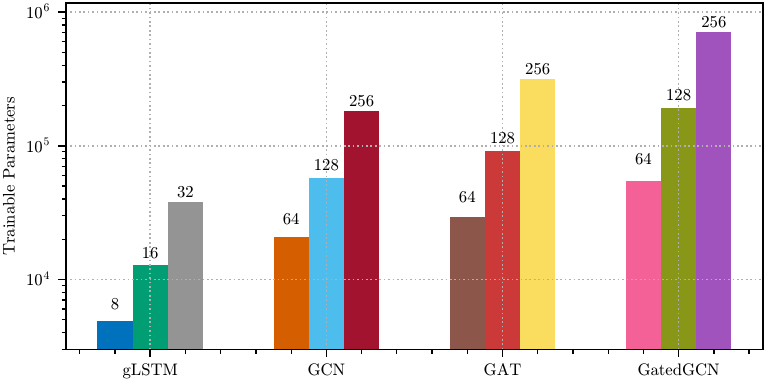}
    \caption{Number of trainable parameters for the expanded set of models tested in \Cref{fig:nar_classification_expanded_performance_no_k_hop,fig:nar_classification_expanded_performance_k_hop}.}
    \label{fig:trainable_parameters_expanded}
\end{figure}

\begin{figure}
    \centering
    \includegraphics[width=0.7\linewidth]{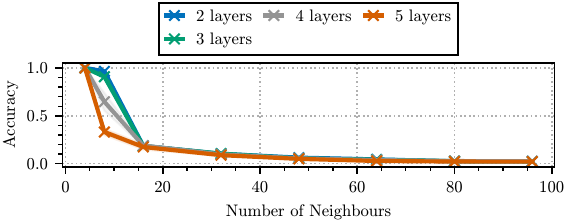}
    \caption{NAR Accuracy for GCN of hidden dimension 128, no K-hop, for varying numbers of GCN layers.}
    \label{fig:variable_depth}
\end{figure}

We additionally verify that the number of layers is not the reason behind GCN being unable to solve NAR at higher neighbor counts. \Cref{fig:variable_depth} visualizes the performance of GCN models with hidden dimension 128 and various layer counts; it transpires that 2 layers performs best out of the tested layer counts.

\subsubsection{Additional Sensitivity Metric Results}

We plot in this section the sensitivity metric trends of \modelname{} vs GCN, both using K-hop aggregation.

\begin{figure}[h]
  \centering
  \setlength{\tabcolsep}{0pt}
  \makebox[\textwidth][c]{%
  \begin{tabular}{cc}
    \subcaptionbox{Jacobian norms\label{fig:nar_jacobian_norms_classification_k_hop}}{
      \includegraphics[height=1.1in]{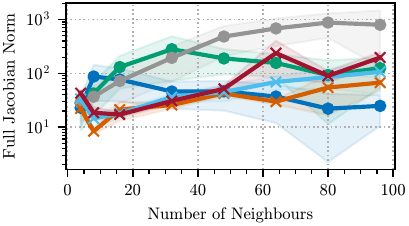}
    } &
    \subcaptionbox{Jacobian norm ratios\label{fig:nar_jacobian_norm_ratios_classification_k_hop}}{
      \includegraphics[height=1.1in]{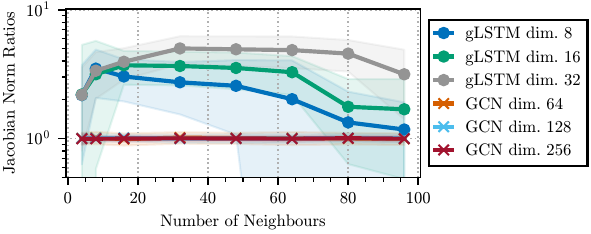}
    }
  \end{tabular}
  }
  \caption{Left: Average Jacobian norms for different \modelname{} and GCN models, with varying number of neighbors in the NAR task. Right: The ratio between the Jacobian norms of the selected (key corresponds to query) to background (key is different from query) neighbor nodes, for the different models - see \Cref{fig:nar_jacobians_separated_classification_k_hop}. This plot differs from that in the main body of the paper in that both \modelname{} and GCN use K-hop aggregation.}
  \label{fig:nar_jacobians_combined_k_hop}
\end{figure}

\begin{figure}[h]
    \centering
    \includegraphics[width=0.8\linewidth]{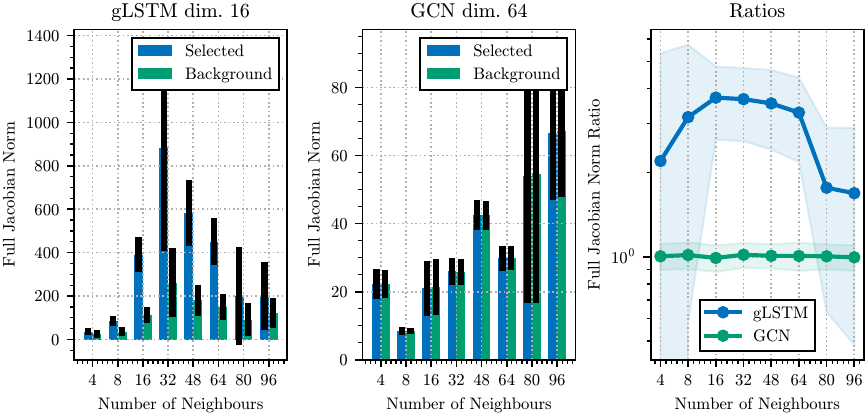}
    \caption{Left: Jacobian norms for \modelname{} of memory dimension 16, with varying number of neighbors in the NAR task, separated by whether the neighbor node corresponds to the given query (selected) or not (background). Middle: Same, with GCN of hidden dimension 64. Right: Ratios of Jacobian norms for selected nodes to background nodes, for these two models. This plot differs from that in the main body of the paper in that both \modelname{} and GCN use K-hop aggregation.}
    \label{fig:nar_jacobians_separated_classification_k_hop}
\end{figure}

\begin{figure}[h]
    \centering
    \includegraphics[width=0.7\textwidth]{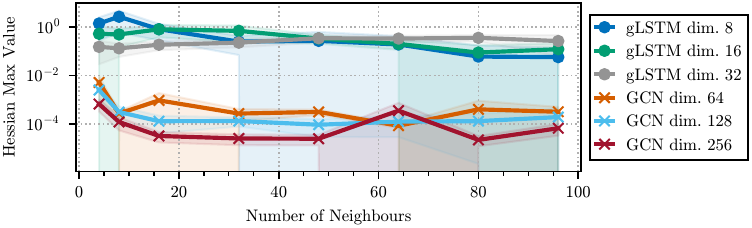}
    \caption{Mean of the maximum Hessian values for different \modelname{} and GCN models, averaged across test set examples and different neighbor nodes. This plot differs from that in the main body of the paper in that both \modelname{} and GCN use K-hop aggregation.}
    \label{fig:nar_hessians_combined_classification_k_hop}
\end{figure}

\Cref{fig:nar_jacobian_norms_classification_k_hop} visualizes the Jacobian norms for different model sizes and numbers of neighbors; \Cref{fig:nar_jacobian_norm_ratios_classification_k_hop} shows the ratios between selected and background node Jacobian norms. \Cref{fig:nar_jacobians_separated_classification_k_hop} separates out the Jacobian norms for \modelname{} memory dimension 16 and GCN hidden dimension 64. \Cref{fig:nar_hessians_combined_classification_k_hop} visualizes the Hessian mixing metric for all models.

\subsection{Neighbor Associative Recall Regression Results}\label{appen:narr}

In this section, we present results for the \emph{regression} variant of the NAR task presented in the main body of the paper. We refer to this as Neighbor Associative Recall Regression (NARR).

Similarly to NAR, for a given neighborhood size $N$ we create a graph of $N +3$ nodes. This graph consists of $N$ ``neighbor'' nodes, a central node to which they are all connected, and an intermediate node connected to the central node and a ``query'' node connected only to the intermediate node.

Each of the neighbor nodes has a feature vector representing a key and a value. The values consist of a fixed-dimensional vector of length $V$ where each element is randomly sampled from a standard normal distribution. The keys are each unique one-hot vectors of dimension $N$. The query node's feature vector contains a single one-hot vector, equal to one of the one-hot vectors of the neighbor nodes. The target of the graph is for the central node to predict the \emph{value} of the neighbor node, corresponding to the key that matches the query node. Each node $u$ is therefore equipped with an input feature vector $\vx \in \mathbb{R}^{V + 2N}$, where the first $V$ elements comprise the value, the next $N$ elements the key and the final $N$ elements the query. Where a node does not have one of these features, the vector elements are set to zero. We note the use of one-hot encoding for keys and values means that the first linear layer of the model acts as a learned embedding function, where multiplication with the one-hot encoding simply selects the corresponding column of the weight matrix. For our experiments, we use $V = 16$.

\begin{figure}
    \centering
    \includegraphics[width=0.6\linewidth]{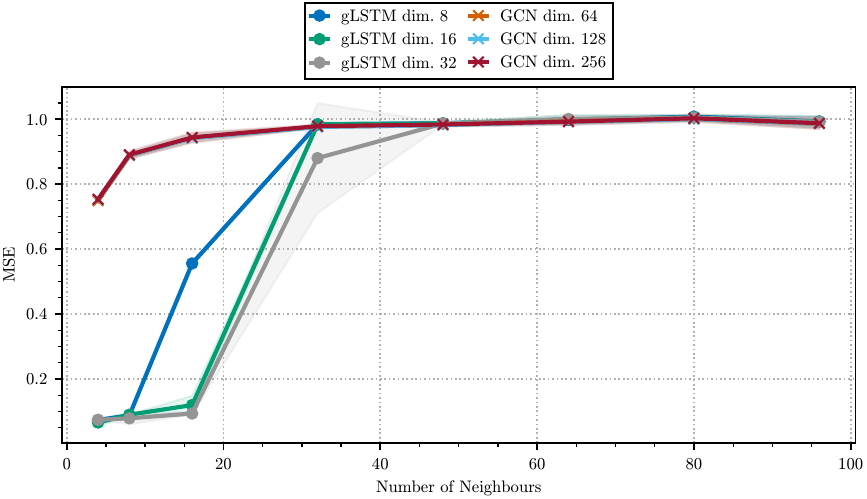}
    \caption{NARR MSE where all models do \emph{not} use K-hop aggregation.}
    \label{fig:nar_regression_expanded_performance_no_k_hop}
\end{figure}

\begin{figure}
    \centering
    \includegraphics[width=0.6\linewidth]{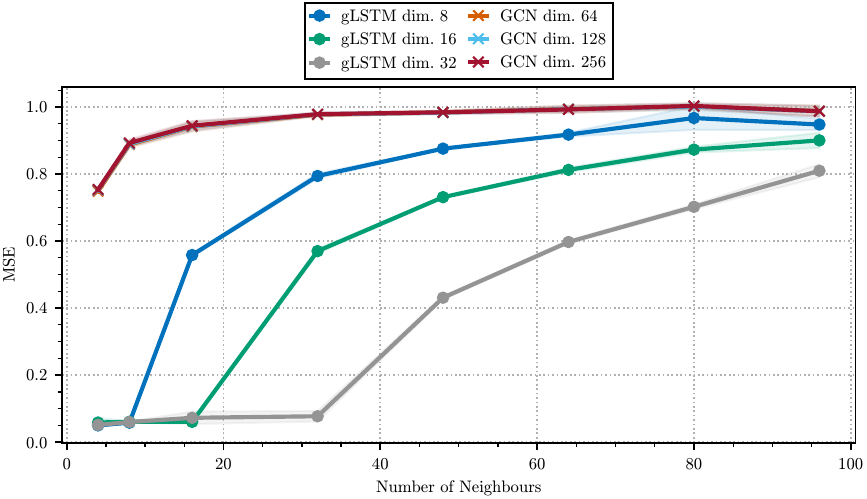}
    \caption{NAR Accuracy where all models \emph{do} use K-hop aggregation.}
    \label{fig:nar_regression_expanded_performance_k_hop}
\end{figure}

Since the value vectors lack the sparsity of NAR, this appears to be a ``harder'' task in the sense that it is more taxing on memory capacity. This means that some of the \osq{} trends are more defined, particularly trends in sensitivity-based measures - see \Cref{appen:narr_sensitivity}. However, our experiments suggest that the regression target means that NARR becomes too hard for vector-memory MPNNs to effectively solve, visible in \Cref{fig:nar_regression_expanded_performance_k_hop,fig:nar_regression_expanded_performance_no_k_hop}.

\begin{figure}
    \centering
    \includegraphics[width=0.5\linewidth]{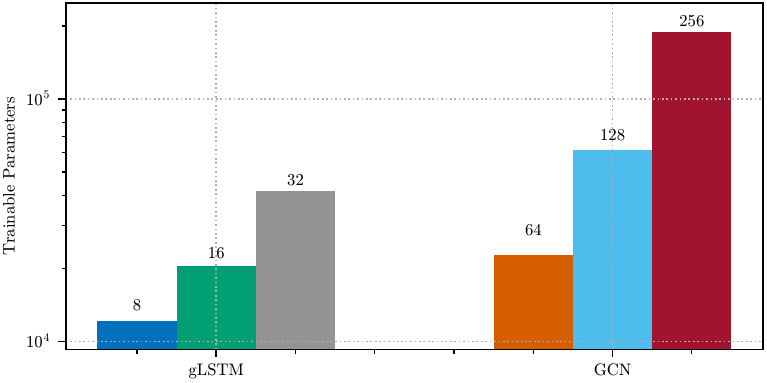}
    \caption{Number of trainable parameters for the expanded set of models tested in \Cref{fig:nar_regression_expanded_performance_no_k_hop,fig:nar_regression_expanded_performance_k_hop}.}
    \label{fig:trainable_parameters_expanded_regression}
\end{figure}

Performance (MSE) for NARR is shown in \Cref{fig:nar_regression_expanded_performance_no_k_hop,fig:nar_regression_expanded_performance_k_hop} for no-K-hop and K-hop aggregation respectively. We note that the performance curves in \Cref{fig:nar_regression_expanded_performance_k_hop} look similar to those obtained by the sequence modeling variant of this experiment in \citet{Schlag2021LinearTA}. The number of trainable parameters is shown in \Cref{fig:trainable_parameters_expanded_regression}.


\subsubsection{Relationship to \OSq{} Sensitivity Metrics}\label{appen:narr_sensitivity}

As with NAR in the main paper, we visualize the behavior of sensitivity-based \osq{} metrics for different neighbor counts and different models. Similarly to the main paper, we compare \modelname{} using K-hop aggregation and GCN without. We note that -- perhaps due to the increased difficulty of the task -- the trends discussed in \Cref{sec:nar_sensitivity} are actually \emph{more} pronounced for the NARR task.

\begin{figure}[h]
  \centering
  \setlength{\tabcolsep}{0pt}
  \makebox[\textwidth][c]{%
  \begin{tabular}{cc}
    \subcaptionbox{Jacobian norms\label{fig:nar_jacobian_norms_regression}}{
      \includegraphics[height=1.1in]{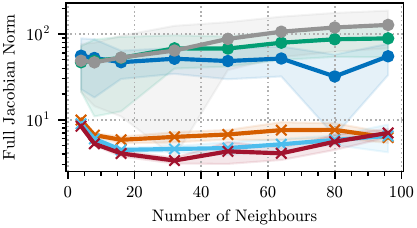}
    } &
    \subcaptionbox{Jacobian norm ratios\label{fig:nar_jacobian_norm_ratios_regression}}{
      \includegraphics[height=1.1in]{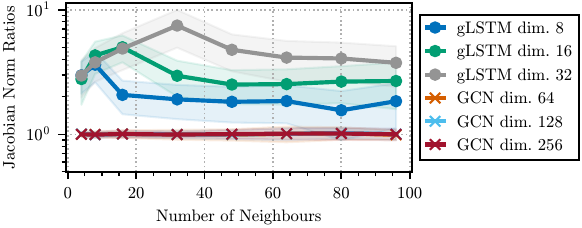}
    }
  \end{tabular}
  }
  \caption{Left: Average Jacobian norms for different \modelname{} and GCN models, with varying number of neighbors in the NARR task. Right: The ratio between the Jacobian norms of the selected (key corresponds to query) to background (key is different from query) neighbor nodes, for the different models - see \Cref{fig:nar_jacobians_separated_regression}.}
  \label{fig:nar_jacobians_combined_regression}
\end{figure}

\begin{figure}[h]
    \centering
    \includegraphics[width=0.8\linewidth]{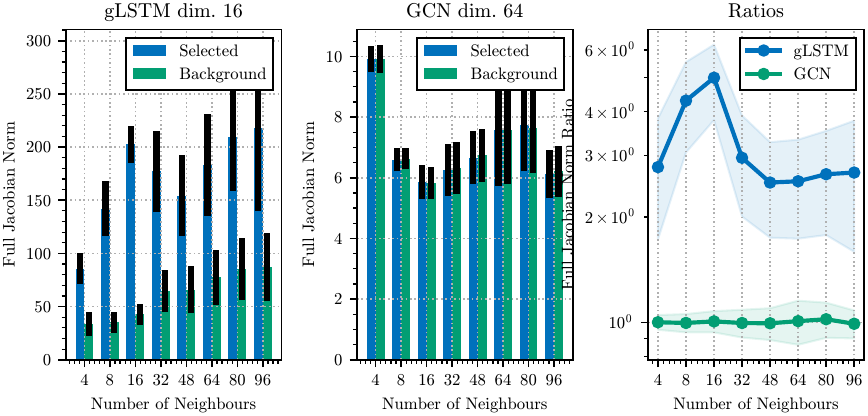}
    \caption{Left: Jacobian norms for \modelname{} of memory dimension 16, with varying number of neighbors in the NARR task, separated by whether the neighbor node corresponds to the given query (selected) or not (background). Middle: Same, with GCN of hidden dimension 64. Right: Ratios of Jacobian norms for selected nodes to background nodes, for these two models.}
    \label{fig:nar_jacobians_separated_regression}
\end{figure}

\begin{figure}[h]
    \centering
    \includegraphics[width=0.7\textwidth]{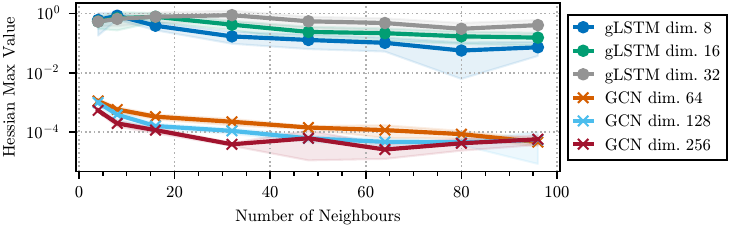}
    \caption{Mean of the maximum Hessian values for different \modelname{} and GCN models, averaged across test set examples and different neighbor nodes.}
    \label{fig:nar_hessians_combined_regression}
\end{figure}

\Cref{fig:nar_jacobian_norms_regression} visualizes the Jacobian norms for different model sizes and numbers of neighbors; \Cref{fig:nar_jacobian_norm_ratios_regression} shows the ratios between selected and background node Jacobian norms. \Cref{fig:nar_jacobians_separated_regression} separates out the Jacobian norms for \modelname{} memory dimension 16 and GCN hidden dimension 64. \Cref{fig:nar_hessians_combined_regression} visualizes the Hessian mixing metric for all models.

We note that the sensitivity difference between selected and background nodes is particularly stark here, even more so for classification-based NAR; \modelname{} consistently shows a sharp drop-off in \Cref{fig:nar_jacobian_norm_ratios_regression} at the memory dimension, and GCN maintains a ratio remarkably close to unity. This closely aligns with the performance of these models, \Cref{fig:nar_regression_expanded_performance_k_hop} demonstrates that \modelname{} performance begins to drop off quickly when the number of neighbors matches the memory dimension, and \Cref{fig:nar_regression_expanded_performance_no_k_hop} demonstrates that GCN is never able to solve the task, for any tested number of neighbors.

We hypothesize that the strong impact of the K-hop aggregation on the model's ability to selectively recall - particularly visible for NARR - may partially explain the dramatic performance decrease when ablating this aggregation, discussed in \Cref{apen:ablations}. We note that, while \modelname{} consistently demonstrates superior performance to GCN, the improved performance is most striking when additionally using K-hop aggregation; it appears that the inductive bias introduced by the K-hop aggregation is particularly suited to the selective recall required by this task.

\end{document}